\documentclass[11pt]{article}
\usepackage[hyperref]{ccl2023-en}
\usepackage{times}
\usepackage{url}
\usepackage{latexsym}
\usepackage{fancyhdr}
\usepackage{multirow}
\usepackage{booktabs}
\usepackage{tabularx}
\usepackage{CJKutf8}
\title{Assessing the Performance of Chinese Open Source Large Language Models in Information Extraction Tasks}

\author{Yida Cai, Hao Sun, Hsiu-Yuan Huang, Yunfang Wu\footnote{Corresponding author.}\\
MOE Key Laboratory of Computational Linguistics, Peking University \\
School of Software and Microelectronics, Peking University\\
School of Computer Science, Peking University\\
{\tt \{caiyida,huang.hsiuyuan\}@stu.pku.edu.cn,\{2301213218,wuyf\}@pku.edu.cn}}
\date{}

\begin{document}
\maketitle
\begin{abstract}
Information Extraction (IE) plays a crucial role in Natural Language Processing (NLP) by extracting structured information from unstructured text, thereby facilitating seamless integration with various real-world applications that rely on structured data. Despite its significance, recent experiments~\cite{zhang2023aligning,wang2023gpt} focusing on English IE tasks have shed light on the challenges faced by Large Language Models (LLMs) in achieving optimal performance, particularly in sub-tasks like Named Entity Recognition (NER). In this paper, we delve into a comprehensive investigation of the performance of mainstream Chinese open-source LLMs in tackling IE tasks, specifically under zero-shot conditions where the models are not fine-tuned for specific tasks. Additionally, we present the outcomes of several few-shot experiments to further gauge the capability of these models. Moreover, our study includes a comparative analysis between these open-source LLMs and ChatGPT\footnote{\url{https://openai.com/blog/chatgpt}}, a widely recognized language model, on IE performance. Through meticulous experimentation and analysis, we aim to provide insights into the strengths, limitations, and potential enhancements of existing Chinese open-source LLMs in the domain of Information Extraction within the context of NLP.

\end{abstract}

\section{Introduction}
\label{intro}

%
%


Information Extraction (IE) is a crucial task in Natural Language Processing (NLP) that involves identifying and extracting relevant information from unstructured text. At its core, IE aims to automatically identify and categorize key elements within text, such as entities, relationships, and events, facilitating a deeper understanding of the textual content\cite{grishman2015information,wei2023zeroshot}. This capability is pivotal for a myriad of applications, from populating knowledge bases and enhancing search engines to powering chatbots and supporting decision-making processes. The essence of IE is to bridge the gap between the richness of human language and the structured data required by computational systems to perform reasoning and analysis.

Generally, IE consists of 3 sub-tasks: Named Entity Recognition (NER), Relation Extraction (RE) and Event Extraction (EE)~\cite{tjong-kim-sang-2002-introduction,ratinov2009design,li2020event}. Among the sub-tasks of IE, NER is perhaps the most fundamental, focusing on the identification of named entities—specifically, categorizing them into predefined groups such as persons, organizations, locations, and dates~\cite{nadeau2007survey,yu2020named}. RE is aimed at identifying semantic relationships between entities within a text~\cite{nasar2021named}. This process involves not only detecting entities but also categorizing the types of relationships between them, such as \textit{is-employed-at} or \textit{is-married-to}. EE focuses on identifying instances of specific events within text and extracting relevant components associated with those events~\cite{liu-etal-2020-event,li2022survey}, which can be defined as an occurrence or an action that happens at a certain time and place, involving participants and entities.

In recent years, Large Language Models (LLMs) have aroused widespread interest and research. LLMs have been proved to achieve impressive performance on various downstream NLP tasks without updating its parameters. In addition, with a few demonstrations added in prompt, LLMs are able to further enhance its performance, which is called In-Context Learning (ICL)~\cite{dong2022survey}. However, experiments~\cite{zhang2023aligning,wei2023zeroshot,wang2023gpt} on English data indicates that LLMs show a relative weak performance on IE tasks such as NER, which is not a difficult task for traditional pre-trained language models like BERT~\cite{devlin2018bert}.

In this paper, we investigate the actual performance of several repesentative Chinese open-source LLMs on Chinese IE tasks, and analyze the performance of different LLMs. In the experiment, we use different methods to obtain a more realistic and comprehensive evaluation. Furthermore, we compare the performance between the existing Chinese open-source LLMs and ChatGPT to verify the gap between them.
\section{Experimental Setup}

\subsection{Methods}
Here we introduce methods for each IE sub-task. Our prompts can be found in Appendix~\ref{app_prompt}

\textbf{NER}: We employ two methods for this task: \textbf{Vanilla} and \textbf{2-Stage}. The \textbf{Vanilla} framework is straightforward, merely providing sentences and a list of entity types, with the expectation that the model will output the extracted entities and their types according to the given format. Apart from that, the \textbf{2-Stage} framework\cite{wei2023zeroshot} divides the task into two parts. In the first step, given a sentence and a list of entity types, the model is asked to output the types of entities that appear in the text. In the second step, each identified entity type is input into the model in turns, prompting the model to output entities corresponding to each type. Compared to the \textbf{Vanilla} framework, the \textbf{2-Stage} framework divides the NER task into two distinct parts. We aim to investigate whether this supplementary round of inference sessions can enhance the NER performance of the Chinese open source LLMs. If proven effective, this would demonstrate that the associated increase in computational cost is justified.

\textbf{RE}: For RE experiments we adopt 2 zero-shot frameworks\cite{Zhang2023LLM-QA4RE}:\textbf{VanillaRE} and \textbf{QA4RE}. In each framework LLMs take sentence, entity1, entity2 and all possible relations as inputs and we use two settings, one with type constraints and the other without. \textbf{VanillaRE} simply lists the sentence, entities and all possible relations for LLM to choose one relation type, whereas \textbf{QA4RE} translate relations into multi-choice linguistic options for LLMs to choose. Unlike the \textbf{VanillaRE} framework, which directly extracts relations, \textbf{QA4RE} transforms the problem into a generative \textbf{Q\&A} format that is better suited for LLMs, whose effectiveness has been proved on English data~\cite{Zhang2023LLM-QA4RE}. We have developed our own verbalization template on a Chinese dataset to convert relation types into answer options, and we aim to assess whether this idea can be generalized to different languages. Compared with RE framework where entities are not available~\cite{wei2023zeroshot}, given entities can help LLMs locate key information and improve performance.

\textbf{EE}: Due to the difficulty of the EE task and the poor experimental performance of the Vanilla framework, in this paper we will only report the results in the \textbf{2-Stage}~\cite{wei2023zeroshot} framework and compare different LLMs' performance. In the first Stage, we ask LLMs to extract all possible event types from several pre-defined event types of the text. After that, LLMs extract argument roles and content for each event type.

\subsection{LLMs}

A total of 5 LLMs are chosen for experiment in this paper: \textbf{ChatGLM3-6B}~\cite{du2022glm}, \textbf{Qwen-7B-Chat} and \textbf{Qwen-14B-Chat}~\cite{qwen}, \textbf{Baichuan2-13B-Chat}~\cite{yang2023baichuan} and \textbf{ChatGPT}\footnote{\url{https://openai.com/blog/chatgpt}}. Except for \textbf{ChatGPT}, all other LLMs are open source and can be deployed on a single GPU, leading to less computility requirement. \textbf{Baichuan2-13B-Chat}, \textbf{Qwen-14B-Chat}, \textbf{ChatGLM3-6B} were recognized for their good performance on the SuperCLUE list\footnote{\url{https://www.superclueai.com/}}~\cite{xu2023superclue}, which were scored with 57.28, 59.73 and 40.32 points respectively. All of our experiments can be performed on 1 A40 GPU.
\subsection{Datasets}
Some details of datasets we use in this paper are shown in Table ~\ref{tab:datasets}. Below are introductions of datasets for each task.

\begin{table}
\centering
\begin{tabular}{c|c|c|c|c}
\toprule
\multicolumn{5}{c}{\textbf{MSRA}}\\
\bottomrule
&\textbf{\#Character} & \textbf{\#LOC} & \textbf{\#ORG} & \textbf{\#PER}\\
\hline
\textbf{Train}&2,171,573&36,860&20,584&17,615\\
\hline
\textbf{Test}&172,601&2,886&1,331&1,973\\
\toprule
\multicolumn{5}{c}{\textbf{Weibo}}\\
\bottomrule
&\textbf{Geo-political} & \textbf{Location} & \textbf{Organization} &\textbf{Person}\\
\hline
\textbf{Named} & 243&88&224&721\\
\hline
\textbf{Nominal} & 0&38&31&636\\
\toprule
\multicolumn{5}{c}{\textbf{DuIE2.0}}\\
\bottomrule
\textbf{\#Instance} & \textbf{\#Entity} & \textbf{\#Relation Type} & \textbf{\#Triple} & \textbf{\#Sentence} \\
\hline
458,184&239,663&49&347,250&214,739\\
\toprule
\multicolumn{5}{c}{\textbf{DuEE}}\\
\bottomrule
&\textbf{\#Document} & \textbf{\#Event} & \textbf{\#Event Type} &  \\
\hline
&11,224&19,640&65&\\
\toprule

\end{tabular}
\caption{Details of datasets we use}
\label{tab:datasets}
\end{table}

\textbf{NER}: MSRA~\cite{levow2006third} and Weibo~\cite{peng2015named} are 2 Chinese NER datasets we choose. MSRA is derived from the newspaper domain, with the goal of identifying persons, locations, and organizations in sentences. Weibo is sourced from the Chinese social media platform Weibo, aiming to identify persons, place names, organizations and geopolitical entities in sentences. 

\textbf{RE}: In this paper we select DuIE2.0~\cite{li2019duie} dataset, which is one of the largest Chinese RE dataset and contains 49 predefined relation types, for this task. Considering the trade-off between experiment and cost, we choose the first 20000 triplets of this dataset for the experiment.

\textbf{EE}: DuEE1.0~\cite{li2020duee} is a Chinese event extraction dataset constracted by Baidu, containing 65 event types.

Since all the dataset was released before 2021, it cannot be ruled out that all current LLMs may have possessed knowledge about these datasets during their training phase. Besides, many LLMs do not disclose information of their training data, so the data contamination issue indeed exists in current Chinese IE benchmark.


\subsection{Evaluation Metrics}

Due to variations in data structures and task requirements, even though the objective of different tasks might be to better align with the standard answers, they will still employ different evaluation methods.

\textbf{NER:} For NER tasks, the primary concern is whether the identified entities and their types match the standard answers. If both the entity name and entity type coincide with the standard answers, the model's prediction is considered correct. The experimental results are evaluated using the micro-F1 score. 

\textbf{RE:} For RE tasks, the output is the relationship between the subject and object. Therefore, a prediction is considered correct only when the predicted relationship is entirely consistent with the true relationship. RE tasks also use the micro-F1 score to evaluate experimental results.

\textbf{EE:} For EE tasks, we use the F1 calculation method~\footnote{\url{https://github.com/PaddlePaddle/PaddleNLP/blob/develop/examples/information_extraction/DuEE/README.md}} provided by DuEE 1.0~\cite{li2020duee}. Specifically, the final F1 score is calculated by measuring the degree of match at the character level.





\section{Main Results}
\subsection{NER Results}

In the Named Entity Recognition (NER) task, we tested both Zero-Shot and Few-Shot approaches. For the Zero-Shot experiments, we utilized 5 LLMs and assessed their performance across 2 datasets. For the Few-Shot experiments, we selected the best-performing open-source model from the Zero-Shot evaluation, Qwen-14B-Chat, to test its performance on 2 datasets.

\subsubsection{Zero-Shot Results}
The Zero-Shot results for NER are shown in the Table~\ref{tab:zero_ner}.We can draw the following conclusions:

\textbf{(1)} Among all LLMs, \textbf{ChatGPT-3.5-Turbo exhibits the best performance}, achieving the highest F1 scores across all datasets and methods. Among all open-source LLMs, Qwen-14B-Chat achieves the highest F1 score on the MSRA dataset, while Baichuan-13B-Chat records the highest F1 score on the Weibo dataset.

\textbf{(2)} Among all open-source LLMs, \textbf{LLM with 13B/14B parameters consistently outperform those with 6B/7B parameters in accuracy across the same methods and datasets}. Except for the Baichuan-13B-Chat, where the 2-Stage method on the Weibo dataset results in an F1 score 0.55 lower than that of Qwen-7B-Chat, all models with 13B/14B parameters demonstrate better F1 scores than those with 6B/7B parameters on the same methods and datasets. The underlying reason is likely due to larger LLMs potentially containing more knowledge relevant to entity recognition, leading to more accurate classification results.

\textbf{(3)} Across 2 datasets and 5 LLMs, totaling 10 experiments, \textbf{the 2-Stage method outperforms the corresponding Vanilla method in terms of F1 score in 7 experiments}. Moreover, in each experiment, the 2-Stage method achieves an accuracy improvement of at least 1.31 and up to 18.01 points over the Vanilla method. This indicates that the 2-Stage method is more effective at focusing the model on identifying entities of the required types, thus enhancing the model's predictive accuracy and, consequently, its F1 score.

\textbf{(4)} Under conditions where the model remains the same, \textbf{the F1 scores on the MSRA dataset are higher than those on the Weibo dataset}. This discrepancy can be attributed to several factors. Firstly, the inherent difficulty level of the two datasets differs. Secondly, the source of the MSRA dataset is newspaper media, which consists of written language, while the Weibo dataset originates from social media, which is more colloquial. Among these, the intrinsic knowledge of the model is more aligned with written language, making it more adept at extracting information from written text.
\begin{table}[t]
\centering

\begin{tabular}{@{}lc|cccccc}
\toprule
 \multicolumn{1}{@{}l}{\multirow{2}{*}{\textbf{Models}}}&\multicolumn{1}{c}{\multirow{2}{*}{\textbf{Methods}}} & \multicolumn{3}{|c}{MSRA} & \multicolumn{3}{c}{Weibo}\\
& &\textbf{P}&\textbf{R}&\textbf{F1}&\textbf{P}&\textbf{R}& \textbf{F1}\\
\midrule
\multirow{2}{*}{ChatGLM3-6B} & Vanilla&11.63&15.90&13.44&6.48&46.5&11.37\\
&2-Stage&15.21&18.98&16.88&7.79&13.63&9.92\\
\midrule
\multirow{2}{*}{Baichuan2-13B-Chat} & Vanilla&15.31&25&18.99&17.11&33.61&\textbf{22.68}\\
&2-Stage&33.32&27.95&31.62&22.5&12.68&16.22\\
\midrule
\multirow{2}{*}{Qwen-7B-Chat} & Vanilla&13.71&27.98&18.4&3.87&21.85&6.57\\
&2-Stage&27.46&28.87&28.15&14.81&20.73&17.05\\
\midrule
\multirow{2}{*}{Qwen-14B-Chat} & Vanilla&34.9&45.57&39.53&20.96&24.36&22.53\\
&2-Stage&50.43&38.89&\textbf{43.91}&29.26&13.45&18.43\\
\midrule
\multirow{2}{*}{ChatGPT-3.5-turbo} & Vanilla&38.94&62.7&47.97&26.67&34.73&30.17\\
&2-Stage&52.44&55.07&\textbf{53.72}&43.55&37.82&\textbf{40.48}\\
\bottomrule
\end{tabular}
\caption{The results of the Zero-Shot experiments conducted on the MSRA and Weibo datasets are presented. \textbf{The bolded values} represent the highest scores among all LLMs and open-source LLMs on the same dataset.}
\label{tab:zero_ner}
\end{table}

\subsubsection{Few-Shot Results}
The Few-Shot results for NER are shown in the Table~\ref{tab:few_ner}. For each method, we conducted Few-Shot experiments with three different shot sizes: 1, 5, and 10. We can draw the following conclusions:

\textbf{(1)}\textbf{The Few-Shot experiments achieved higher F1 scores than Zero-Shot in 3/4 of the cases}, indicating that models improve their understanding of tasks when provided with examples. However, in the 2-Stage method, except for an increase in F1 from Zero-Shot to One-Shot, increasing the number of shots did not significantly change or even decreased the F1 score, primarily due to a substantial drop in precision. This might be because the 2-Stage method decomposes the NER task into two sub-tasks. With more examples, the model tends to generate more types in the first task to fit the diverse types in the examples, leading to the model likely outputting a potential answer in the second task when asked about a non-existent type, thereby significantly reducing precision and further decreasing the F1 score.

\textbf{(2)} In the Vanilla method, except for a decrease in precision when increasing from 1 shot to 5 shots on the Weibo dataset, \textbf{all other experiments showed that precision on the same dataset increases with the number of shots}. This is because the Vanilla method's prompt lacks detailed task description, leading to a model's vague understanding of the types of entities to predict in the Zero-Shot experiment. After examples are provided, the model gains a more accurate understanding of which entities to extract, significantly increasing the precision from Zero-Shot to One-Shot. Subsequently, by increasing the number of shots, the model can understand the task more accurately, further enhancing precision.

\textbf{(3)} \textbf{In the 2-Stage method, the experimental results showed a different pattern of change}. Except for a slight decrease in recall when increasing from 5 shots to 10 shots on the Weibo dataset, all other experiments demonstrate that recall on the same dataset would increase with the number of shots. This phenomenon might be attributed to the point mentioned in (1), which suggests that when more content is generated, there's also a certain degree of improvement in the experiment's recall.
\begin{table}[t]
\centering
\begin{tabular}{@{}ll|c|cccccc}
\toprule
 \multicolumn{1}{@{}l}{\multirow{2}{*}{\textbf{Models}}}&\multicolumn{1}{@{}c}{\multirow{2}{*}{\textbf{Methods}}} &\multicolumn{1}{|c}{\multirow{2}{*}{\textbf{Shot-Num}}}& \multicolumn{3}{|c}{MSRA} & \multicolumn{3}{c}{Weibo}\\
&& &\textbf{P}&\textbf{R}&\textbf{F1}&\textbf{P}&\textbf{R}& \textbf{F1}\\
\midrule
\multirow{8}{*}{Qwen-14B-Chat} & \multirow{4}{*}{Vanilla}
& 0shot&34.9&45.57&39.53&20.96&24.36&22.53\\
&& 1shot&53.01&63.83&\textbf{57.92}&39.13&20.17&26.62\\
&& 5shot&60.33&42.21&49.67&36.67&23.33&28.52\\
&& 10shot&65.72&43.66&52.46&48.64&21.8&\textbf{30.13}\\
\cmidrule{3-9}
&\multirow{4}{*}{2-Stage}
& 0shot&50.43&38.89&43.91&29.26&13.45&18.43\\
&& 1shot&48.94&52.29&\textbf{50.56}&30.05&33.33&\textbf{31.61}\\
&& 5shot&32.28&59.25&41.79&27.91&34.84&30.99\\
&& 10shot&34.25&60.05&43.62&28.57&32.12&30.24\\

\bottomrule
\end{tabular}
\caption{The results of the Few-Shot experiments conducted on the MSRA and Weibo datasets are presented. \textbf{The bolded values} represent the highest scores among experiments of the specific method.}
\label{tab:few_ner}
\end{table}

\subsection{RE results}
The main result of Chinese Zero-Shot RE are shown in Table~\ref{tab:zero_re}. We have the following 4 observations based on our experiments. 

\textbf{(1) Most LLMs perform well with type constraints, but the performance drops significantly without type constraints.} Without type constraints, each LLMs needs to choose one from 50 relation types, and the constraints reduce a lot of interference information for LLMs. Notably, on some entity pairs of types only correct relation and NoTA(none-of-the-above) relation are retained. 

\textbf{(2) Performances on traditional Vanilla framework are better than those on QA4RE framework.} On the one hand, it may be that the RE task of DuIE2.0 dataset is not very difficult for LLMs, and the introduction of QA has increased the difficulty of understanding. On the other hand, in the no constraints experiment, many unreasonable answer options are added to prompt (such as \textit{Alice is the publisher of Bob}). These information interferes with the selection of relation types for LLMs.

\textbf{(3) LLMs show similar performances in 2 settings.} ChatGLM3-6B, Qwen-7B-Chat, Qwen-14B-Chat and ChatGPT-3.5-turbo all show a leading performance in both type constrained and unconstrained settings. Among all open source LLMs, Qwen-14B-Chat has the best performance, only 0.03 percentage lower than ChatGPT-3.5-turbo in type constrained setting.

\textbf{(4) The performance of ChatGPT-3.5-turbo in the unconstrained QA4RE setting far exceeds other LLMs (70.01 vs. 47.35)}, indicating that the anti-interference ability of ChatGPT is much stronger than all open-source LLMs when prompt is relatively long and consists of unreasonable sentences. However, considering the comparison of model sizes, such difference is actually acceptable (175B vs. the largest 14B).

When discussing the limitations of the DuIE 2.0 dataset, it is important to note that in type-constrained settings, many entity pairs retain only the correct relation type and NoTA (None of The Above) relation. For example, the possible relation types between \textit{publishing house} and \textit{book} are limited to \textit{publication} or \textit{NoTA}. This partly explains why LLMs perform exceptionally well with type constraints.
\begin{table}[tp]
\centering

\begin{tabular}{@{}ll|cccccc}
\toprule
 \multicolumn{2}{@{}l}{\multirow{2}{*}{\textbf{Models}}} & \multicolumn{3}{|c}{type constrained} & \multicolumn{3}{c}{no constraint}\\
& &\textbf{P}&\textbf{R}&\textbf{F1}&\textbf{P}&\textbf{R}& \textbf{F1}\\
\midrule
\multirow{2}{*}{Chatglm3-6B} & VanillaRE&89.48&89.78&89.63&73.61&73.6&73.60\\
&QA4RE&88.30&88.04&88.17&16.55&16.49&16.52\\
\midrule
\multirow{2}{*}{Baichuan2-13B-Chat} & VanillaRE&82.05&38.72&52.61&65.95&48.80&56.09\\
&QA4RE&81.02&80.95&80.98&10.32&10.32&10.32\\
\midrule
\multirow{2}{*}{Qwen-7B-Chat} & VanillaRE&91.81&90.74&91.27&75.817&75.819&75.815\\
&QA4RE&87.64&85.27&86.44&4.93&4.55&4.73\\
\midrule
\multirow{2}{*}{Qwen-14B-Chat} & VanillaRE&89.62&87.46&88.53&78.53&76.90&\textbf{77.70}\\
&QA4RE&95.84&91.12&\textbf{93.42}&45.57&49.28&47.35\\
\midrule
\multirow{2}{*}{ChatGPT-3.5-turbo} & VanillaRE&94.65&92.28&\textbf{93.45}&88.17&81.41&\textbf{84.66}\\
&QA4RE&94.41&87.92&91.05&70.35&69.68&70.01\\
\bottomrule
\end{tabular}
\caption{Experimental result of zero-shot RE on a 20000 subset of DuIE2.0 dataset(\%). For each LLM, we mark the better result in \textbf{bold} to compare the two framework.}
\label{tab:zero_re}
\end{table}

\subsection{EE results}
\begin{table}[t]
\centering

\begin{tabular}{@{}l|ccc}
\toprule
 \multicolumn{1}{@{}l}{\multirow{2}{*}{\textbf{Models}}} & \multicolumn{3}{|c}{DuEE1.0} \\ &\textbf{P}&\textbf{R}&\textbf{F1}\\
\midrule
\multirow{1}{*}{ChatGLM3-6B} & 16.04&44.29&23.55\\
\midrule
\multirow{1}{*}{Baichuan2-13B-Chat} & 32.32&51.35&39.68\\
\midrule
\multirow{1}{*}{Qwen-7B-Chat} & 27.59&22.43&24.75\\
\midrule
\multirow{1}{*}{Qwen-14B-Chat} & 38.87&33.59&36.03\\
\midrule
\multirow{1}{*}{ChatGPT-3.5-turbo~\cite{wei2023zeroshot}} & 74.6&67.5&\textbf{70.9}\\
\bottomrule
\end{tabular}
\caption{Results of the Zero-Shot EE experiments.The \textbf{bold} values represent the highest scores among all LLMs.}
\label{tab:zero_ee}
\end{table}

The Zero-Shot results for EE are presented in Table~\ref{tab:zero_ee}. Given the higher demand of the EE task on model's reading comprehension ability, only the 2-Stage decomposition method enables the production of clearer answers. ChatGPT continues to lead with the highest F1 score among all LLMs, largely due to its superior ability to grasp the complex output format requirements in the prompt. Among open-source models, Baichuan2-13B-Chat demonstrates the best performance; however, its F1 score still remains 31.22 points lower than that of ChatGPT.



\section{Related Work}
\subsection{Chinese Information Extraction}

In recent years, thanks to the development of deep learning, Information Extraction (IE) tasks have evolved from rule-based and statistical machine learning~\cite{li2004svm} approaches to those based on deep learning~\cite{yang2022survey}, particularly in the direction of Transformer~\cite{han2021transformer}.Chinese Information Extraction tasks, as an essential branch of Information Extraction tasks, have also experienced significant development with the aid of Transformers.~\cite{li2019dice} improved the performance of NER tasks by designing a loss function on a BERT-based NER model.~\cite{cohen2020relation} and ~\cite{wang2022deepstruct} also achieved good results on RE and EE tasks, respectively, by improving the Transformer model.

However, high-quality labeled data is always hard to obtain. Aside from the success of these supervised learning models, researchers in the field have also turned their attention to Zero/Few-Shot learning research, such as Few-Shot Relation Extraction~\cite{sainz-etal-2021-label} and Few-Shot Event Extraction~\cite{sainz-etal-2022-textual}.

\subsection{LLMs on IE}

Due to ChatGPT's notable performance in recent years, many IE tasks have been conducted on ChatGPT~\cite{wang2023gpt,wei2023zeroshot}.~\cite{wei2023zeroshot} introduced a novel two-stage paradigm for Information Extraction (IE) involving interactive communication. Initially, the process involves prompting ChatGPT to identify the categories of elements present in the text. Following this, the second stage entails requesting ChatGPT to pinpoint and retrieve the specific instances of each previously identified category.~\cite{wang2023gpt} employed in-context learning (ICL) for Named Entity Recognition (NER) by integrating specific tokens into examples drawn from the training dataset.Unlike previous work, we aim to apply methods used on ChatGPT to a broader range of LLMs, further exploring the performance of open-source LLMs in Chinese Information Extraction tasks under Zero/Few-Shot scenarios.

\section{Conclusion}
In this work, we begin by asserting the observation that English Large Language Models (LLMs) exhibit subpar performance in Information Extraction (IE) tasks. Building upon this premise, we proceed to evaluate the performance of several Chinese open-source LLMs alongside ChatGPT, aiming to gauge their effectiveness in IE tasks. Our experimental approach primarily focuses on zero-shot learning across three core IE sub-tasks: Named Entity Recognition (NER), Relation Extraction (RE), and Event Extraction (EE).

Our findings reveal that ChatGPT surpasses all Chinese open-source LLMs across all IE tasks, showcasing superior performance. While the Qwen series and Baichuan LLMs demonstrate relatively commendable performance in NER tasks, they still fall short compared to the ChatGPT series. On the other hand, the Qwen series LLMs demonstrate stronger performance in RE tasks, nearing the level of ChatGPT, yet all open-source LLMs lag notably behind ChatGPT in EE tasks.

In our future endeavors, we plan to delve deeper into the performance of Chinese LLMs under Few-shot or fine-tuning settings for IE sub-tasks. Furthermore, we aim to explore innovative model architectures to augment the performance of LLMs specifically tailored for Chinese IE tasks. These endeavors are expected to contribute significantly to advancing the capabilities of Chinese LLMs in the realm of Information Extraction.


\bibliographystyle{ccl}
\bibliography{ccl2023-en}
\appendix
\section{Prompts for IE}
\label{app_prompt}

\begin{table*}[h]
\centering
\begin{tabularx}{\textwidth}{X|X}

\toprule
\multicolumn{2}{c}{Prompt for NER}\\
\bottomrule
\multicolumn{1}{c}{Vanilla} & \multicolumn{1}{c}{2-Stage} \\
\bottomrule
\begin{CJK}{UTF8}{gbsn}
假设你是一个命名实体识别模型，接下来我会给你一个句子，请根据我的要求找出每个句子中的实体并给出它的实体类型，实体类型只有：人物、地点、机构，没有其他类型。\newline
请按照以下的格式输出：["实体名称1","实体类型1"], ["实体名称2","实体类型2"], …。如果该句子中不含有指定的实体类型，你可以输出:[]。除了这些以外请不要输出别的多余的话,例如:"这个是输出的结果"。
\end{CJK}
& 
\begin{CJK}{UTF8}{gbsn}
\textbf{Stage 1}: 请输出句子中出现过的实体类型，实体类型只有三种：机构、地点、人物，没有其他类型。请将句子中存在的实体类型名称用','隔开输出。
\newline
\textbf{Stage 2}: 句子中对应"{}"类型的实体名称有哪些，请按照以下的格式输出：["实体名称1","实体类型1"], ["实体名称2","实体类型2"], …。
\end{CJK}
\\
\toprule

\end{tabularx}
\caption{Prompt for NER experiments}
\label{tab:prompt_ner}
\end{table*}
\begin{table*}[h]
\centering
\begin{tabularx}{\textwidth}{X}
\toprule
\multicolumn{1}{c}{2-Stage Prompt for EE}\\
\bottomrule
\begin{CJK}{UTF8}{gbsn}
\textbf{Stage 1}:
给定的句子为："\{句子\}"\newline
给定事件类型列表：\{类型列表\}\newline
在这个句子中，可能包含了上述列表中的哪些事件类型？\newline
请给出事件类型列表中的事件类型。\newline
如果不存在则回答：无\newline
按照元组形式回复，如 (事件类型1, 事件类型2, ……)：\newline
\textbf{Stage 2}:
事件类型"\{\textit{类型}\}"对应的论元角色列表为：\{\textit{论元列表}\}。\newline
在"\{\textit{句子}\}"这句话中，请提取出上述每一个论元角色对应的事件论元内容。\newline
如果论元角色没有相应的论元内容，则论元内容回答：无\newline
请将结果按照以下的列表形式回复：[论元角色1,论元内容1],[论元角色2,论元内容2]...。\newline不要说无关的话例如：“这是找到的结果”：
\end{CJK}
\\
\toprule
\end{tabularx}
\caption{Prompt for EE experiments}
\label{tab:prompt_ee}
\end{table*}
\begin{table*}[tp]
\centering
\begin{tabularx}{\textwidth}{X|X}

\toprule
\multicolumn{2}{c}{Prompt for RE (No Constraints)}\\
\bottomrule
\multicolumn{1}{c}{VanillaRE} & \multicolumn{1}{c}{QA4RE} \\
\bottomrule
\begin{CJK}{UTF8}{gbsn}
给定一个句子和句子中的两个实体，根据所提供的句子对这两个实体之间的关系进行分类。下面列出了所有可能的关系:\newline
- 主演\newline
- 目\newline
- 身高\newline
- 出生日期\newline
- 国籍\newline
- 连载网站\newline
- 作者\newline
- 歌手\newline
- 海拔\newline
- 出生地\newline
- 导演\newline
- 气候\newline
- 朝代\newline
- 妻子\newline
- 丈夫\newline
- 民族\newline
- 毕业院校\newline
- 编剧\newline
- 无关系\newline
句子: 如何演好自己的角色，请读《演员自我修养》《喜剧之王》周星驰崛起于穷困潦倒之中的独门秘笈\newline
实体1: 喜剧之王\newline
实体2: 周星驰\newline
关系:
\end{CJK}
& 
\begin{CJK}{UTF8}{gbsn}
请问可以从给定的句子中推断出以下哪个选项。\newline
句子: 如何演好自己的角色，请读《演员自我修养》《[喜剧之王]》[周星驰]崛起于穷困潦倒之中的独门秘笈\newline
Options:\newline
1. [喜剧之王]的主演是[周星驰]\newline
2. [喜剧之王]在分类学上属于[周星驰]\newline
3. [喜剧之王]的身高是[周星驰]\newline
4. [喜剧之王]的出生日期是[周星驰]\newline
5. [喜剧之王]的国籍是[周星驰]\newline
6. [喜剧之王]连载于[周星驰]\newline
7. [喜剧之王]的作者是[周星驰]\newline
8. [喜剧之王]的歌手是[周星驰]\newline
9. [喜剧之王]的海拔是[周星驰]\newline
10. [喜剧之王]的出生地是[周星驰]\newline
11. [喜剧之王]的导演是[周星驰]\newline
12. [喜剧之王]的气候是[周星驰]\newline
13. [喜剧之王]的朝代是[周星驰]\newline
14. [喜剧之王]的妻子是[周星驰]\newline
15. [喜剧之王]的丈夫是[周星驰]\newline
16. [喜剧之王]的民族是[周星驰]\newline
17. [喜剧之王]的毕业院校是[周星驰]\newline
18. [喜剧之王]的编剧是[周星驰]\newline
19. [喜剧之王]和[周星驰]无上述关系\newline
从给定的句子中可以推断出以下哪个选项?\newline
选项:
\end{CJK}
\\
\toprule

\end{tabularx}
\caption{Prompt for RE experiments. We have \textbf{omitted some options} here, and each prompt in the experiment includes 50 options}
\label{tab:prompt_re}
\end{table*}
\end{document}